\documentclass[runningheads]{llncs}

\usepackage{eccv}

\usepackage{eccvabbrv}
\usepackage[pagebackref,breaklinks,colorlinks,citecolor=eccvblue]{hyperref}
\usepackage{graphicx}
\usepackage{booktabs}
\usepackage{wrapfig}

\usepackage[accsupp]{axessibility}  

\usepackage{hyperref}

\usepackage{orcidlink}
\usepackage{arydshln}
\usepackage[table]{xcolor}
\usepackage{multirow}
\begin{document}

\title{RT-SDGOD: Real-Time Single-Domain Generalized Object Detection} 

\titlerunning{Abbreviated paper title}

\author{Yupeng Zhang\inst{1,2} \and
Fangzhuo Gao\inst{1,2} \and
Ruize Han\inst{3} \and
Wei Feng\inst{1,2} \and
\\Liang Wan\inst{1,2}}


\institute{College of Intelligence and Computing, Tianjin University \and
Key Research Center for Surface Monitoring and Analysis of Relics, State Administration of Cultural Heritage \and
Faculty of Computer Science and Artificial Intelligence, Shenzhen University of Advanced Technology \\
\email{\{zhangyupeng, fangzhuo, wfeng, lwan\}@tju.edu.cn, hanruize@suat-sz.edu.cn}}


\maketitle

\begin{abstract}
In real-world deployment under strict real-time constraints, weather and imaging variations induce significant distribution shifts, severely degrading detectors.
Single-Domain Generalized Object Detection aims to mitigate this issue, yet existing methods rarely investigate—at the level of problem formulation—the generalization capability of real-time detectors under such constrained inference budgets.
To this end, we introduce Real-Time Single-Domain Generalized Object Detection (RT-SDGOD), which focuses on how real-time detectors can achieve cross-domain generalization under zero extra inference overhead by relying solely on training-time representation learning.
We observe that, under domain shift, DETR-based real-time detectors mainly degrade through increased missed detections, rooted in limited and unstable object-level discriminative evidence.
Based on this, we propose RT-SDGDet, a multi-evidence collaborative modeling framework for RT-SDGOD. 
The core idea is to enable multiple queries of the same object to collaboratively cover more sufficient discriminative evidence while maintaining the stability of such evidence modeling across views. 
Specifically, we use one-to-many (O2M) supervision to construct stable object-specific query groups, and further design Discriminative Evidence Diversity Learning (DEDL) and Dual-view Evidence Consistency Learning (DvECL) to expand object-level evidence coverage and improve evidence stability under appearance perturbations, respectively. 
Since all components are introduced only during training, our method incurs no extra inference overhead. 
Extensive experiments show that the proposed method achieves better generalization performance than existing approaches across multiple unseen target domains.

\end{abstract}

\section{Introduction}
\label{sec:intro}
In latency-sensitive real-world vision systems such as autonomous driving and intelligent surveillance, object detectors must achieve high accuracy under strict latency, model size, and computational constraints. 
To meet this demand, recent real-time detectors have continuously improved the trade-off between inference speed and accuracy, achieving strong progress.
It should not be overlooked that real-world changes in illumination, weather, and imaging quality often introduce substantial distribution shifts, severely degrading detection performance.

\begin{figure}[t!]
	\centering
	\includegraphics[width=1.0\linewidth]{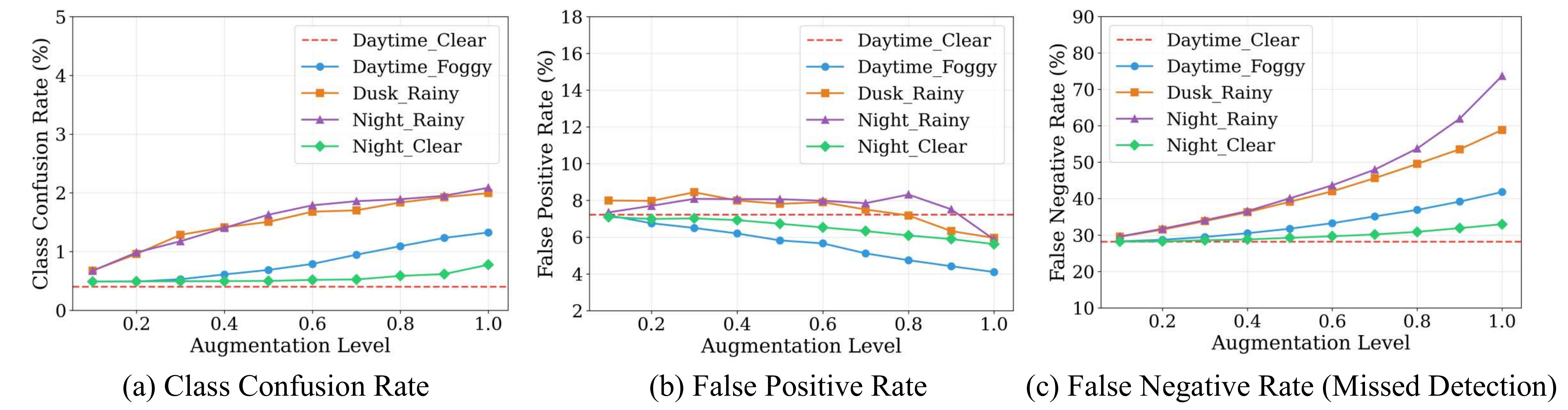}
	\vspace{-20pt}
    	\caption{Error pattern analysis of DETR-based detectors under SDGOD. Taking predictions on daytime-clear source images as the reference, we progressively increase environmental perturbations to simulate increasingly severe domain shifts and analyze the resulting error trends. (a) Class Confusion Rate: the proportion of detected objects with incorrect categories; (b) False Positive Rate: the proportion of detections unmatched to any GT object; (c) False Negative Rate: the proportion of missed GT objects. As domain shift increases, class confusion remains low with only a slight rise, false positives decrease marginally, while false negatives grow steadily and become the dominant source of performance degradation.}  
	\label{fig:moti}
	\vspace{-20pt}
\end{figure}

Single-Domain Generalized Object Detection (SDGOD) aims to train a detector using labeled data from a single source domain and generalize it stably across multiple unseen target domains. Existing studies have explored this problem through data augmentation (\eg, CLIP the Gap~\cite{vidit2023clip}), feature disentanglement (\eg, DG-DETR~\cite{hwang2025dg}), architecture search (\eg, G-NAS~\cite{wu2024g}), and test-time adaptation (\eg, SA-DETR~\cite{han2025style}). Although effective in offline evaluation, most of these methods are not designed for strict real-time deployment, as they are often built on two-stage frameworks such as Faster R-CNN or rely on extra auxiliary modules and test-time mechanisms. As a result, existing studies rarely formulate the generalization ability of real-time detectors under tightly constrained inference budgets as an explicit problem setting.

To address this gap, we further introduce a stricter and more deployment-faithful setting, termed \textbf{Real-Time Single-Domain Generalized Object Detection (RT-SDGOD)}. In this setting, the detector is trained only on labeled data from a single source domain and evaluated on multiple unseen target domains, while the test-time inference pipeline must remain strictly unchanged, \ie, no additional computation, auxiliary modules, or adaptation mechanisms are allowed. As a result, cross-domain robustness must come entirely from training-time representation learning. RT-SDGOD therefore \textit{shifts the focus from test-time compensation to whether the detector itself can learn stable and transferable object representations} under zero extra inference overhead.

Recently, DETR-based real-time detectors, which are built on set prediction, enable end-to-end detection, and exhibit a strong speed–accuracy trade-off, making them an attractive foundation for RT-SDGOD.
However, our systematic experiments reveal a stable and representative failure pattern of these detectors under the SDGOD setting. Specifically, when weather and illumination perturbations are progressively intensified under controlled conditions to simulate domain shifts of increasing severity, the resulting performance drop is not mainly characterized by increased class confusion. As shown in Fig.~\ref{fig:moti}, class-confusion errors remain largely stable, and false positives even decrease as the domain shift becomes stronger, whereas false negatives consistently increase and become the dominant source of performance degradation.

\begin{wrapfigure}{r}{0.5\textwidth} \vspace{-23pt}
	\centering
	\includegraphics[width=1.0\linewidth]{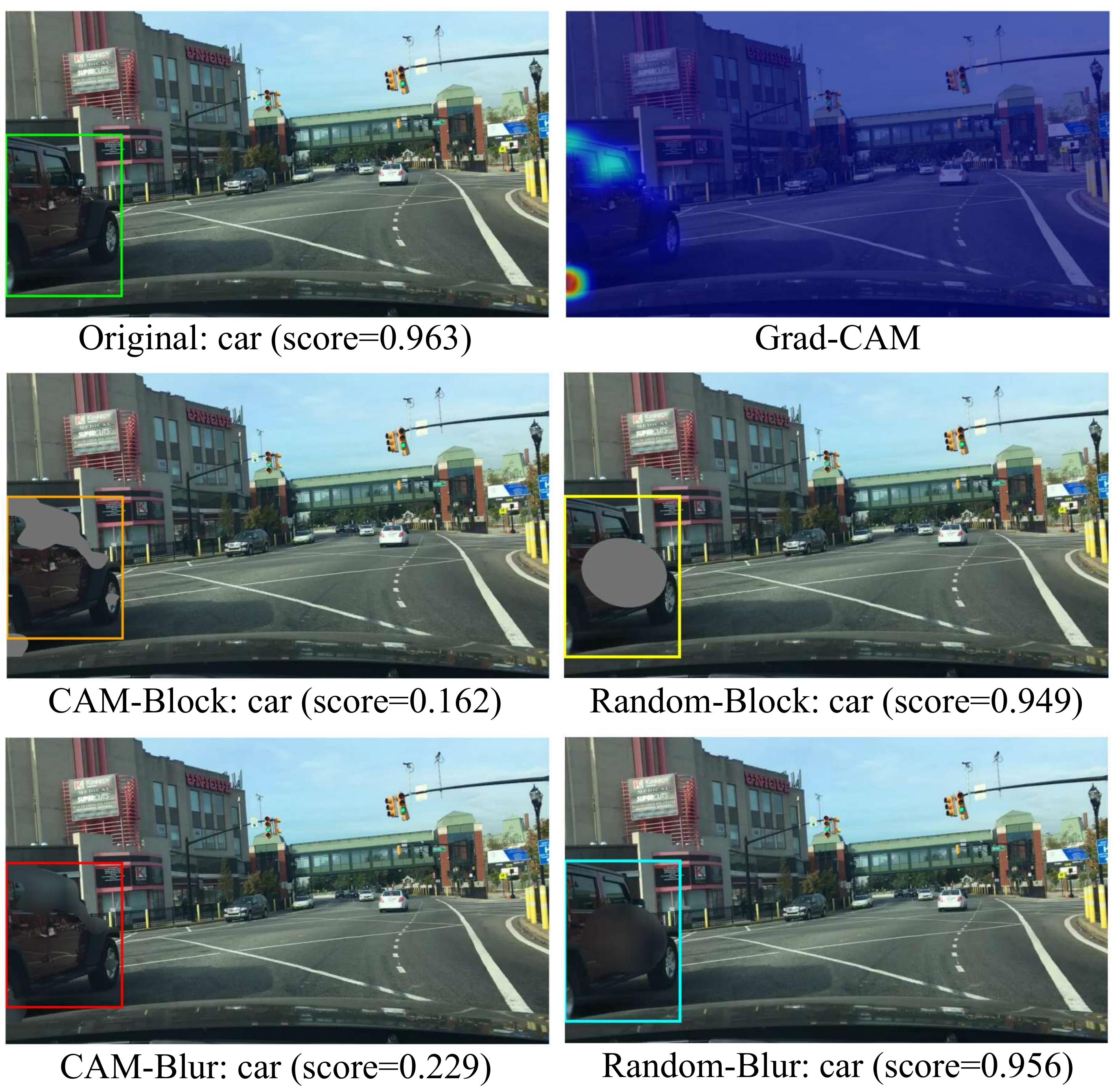}
	\vspace{-20pt}
	\caption{Evidence concentration analysis of DETR-based real-time detectors. For a high-confidence prediction, we first visualize the detector response with Grad-CAM and further examine its sensitivity to different regions through local blocking and local blurring. Specifically, CAM-Block and CAM-Blur apply blocking and blurring to the Grad-CAM-highlighted high-response region, while Random-Block and Random-Blur apply the same perturbations to random regions of similar size.} 
	\label{fig:hot}
	\vspace{-20pt}
\end{wrapfigure}
Further analysis shows that this performance degradation fundamentally stems from the limitedness and instability of object-level discriminative evidence. Through detection-response visualization and local occlusion/blurring experiments (Fig.~\ref{fig:hot}), we find that high-confidence predictions of DETR-based real-time detectors often rely heavily on a few local discriminative cues rather than the broader appearance information of the object. Here, local occlusion simulates the loss of critical details under extreme weather, while local blurring corresponds to local representation shifts caused by imaging changes. The results show that applying the same perturbation to non-critical regions leaves detection scores almost unchanged, whereas disturbing these high-response regions causes a sharp drop in confidence. This indicates that existing detectors make decisions based on a small set of fragile local cues, leading to insufficient evidence coverage and rapid instability under domain shift when key evidence deteriorates or object representations shift.

The above analysis indicates that the core challenge of RT-SDGOD is how to, during training, mitigate existing detectors’ over-reliance on a few concentrated local cues and suppress the evidence distribution shift induced by domain change, thereby learning more sufficient and stable discriminative evidence so that the model can maintain stable object-level activations and matching even when key local evidence degrades or representations drift.

To this end, we propose RT-SDGDet, a multi-evidence collaborative modeling framework for RT-SDGOD. Built on the DETR-based real-time detector RF-DETR, our method explicitly enhances the sufficiency and stability of object-level discriminative evidence during training without changing the test-time inference pipeline or overhead. Its core idea is to enable multiple queries of the same object to collaboratively cover more sufficient discriminative evidence while maintaining the stability of such evidence modeling across views. Specifically, we first use one-to-many (O2M) supervision to construct stable object-specific query groups as the basis for multi-query collaboration within the same object. 
We then introduce \textbf{Discriminative Evidence Diversity Learning (DEDL)}, which constructs explicit discriminative evidence descriptors from query-wise deformable cross-attention and encourages queries within the same group to attend to complementary discriminative evidence through diversity regularization, thereby expanding object-level evidence coverage.
Furthermore, we introduce \textbf{Dual-view Evidence Consistency Learning (DvECL)} to align cross-view queries that belong to the same object and rely on similar discriminative evidence, improving the stability of discriminative evidence under appearance perturbations. Since all components are introduced only during training, our method improves cross-domain robustness without any additional inference overhead. 
Extensive experiments on multiple unseen target domains show that our method achieves better generalization performance than existing approaches.




\section{Related Work}
\textbf{Real-Time Detector.}
Real-time object detection has long been dominated by one-stage detectors, from early YOLO~\cite{redmon2016you} to more recent variants such as YOLOv12~\cite{tian2025yolov12} and v13~\cite{lei2025yolov13}, which continuously improve the accuracy-speed trade-off. In parallel, DETR~\cite{carion2020end} reformulates detection as set prediction and removes hand-crafted components such as anchors and non-maximum suppression through an end-to-end Transformer-based framework. Building on this paradigm, a series of real-time DETR detectors have recently emerged, including RT-DETR v1-v4~\cite{zhao2024detrs,lv2024rt,wang2025rt,liao2025rt}, LW-DETR~\cite{chen2024lw}, D-FINE~\cite{peng2024d}, DEIM v1-v2~\cite{huang2025deim,huang2025real}, and RF-DETR~\cite{robinson2025rf}. These advances make DETR-based real-time detectors increasingly competitive under strict latency constraints, and thus a natural foundation for studying RT-SDGOD. Our work is built upon RF-DETR and focuses on improving cross-domain generalization without introducing any extra inference overhead.

\textbf{Single-Domain Generalized Object Detection (SDGOD).} 
Existing SDGOD methods can be broadly grouped into four categories. Data augmentation methods improve robustness by perturbing images or features to enlarge the training distribution, such as CLIP the Gap~\cite{vidit2023clip}, DivAlign~\cite{danish2024improving}, and SRCD~\cite{rao2024srcd}; however, most still rely on two-stage detectors or additional modules. Feature disentanglement methods separate domain-invariant and domain-specific factors through dedicated architectures or loss designs, as in SDGOD~\cite{wu2022single}, DG-DETR~\cite{hwang2025dg}, and UFR~\cite{liu2024unbiased}, but they usually introduce extra feature computation or inference branches. Architecture search methods, such as G-NAS~\cite{wu2024g}, seek more generalization-friendly detector architectures, yet their inference efficiency depends on the searched structure. Test-time adaptation methods, \eg, SA-DETR~\cite{han2025style}, adapt the model to unseen domains through dynamic inference-time adjustments, but inevitably incur extra test-time overhead. \textbf{Notably, the only two existing DETR-based methods also fail to balance these two aspects well.} In contrast, our method is built on a DETR-based real-time detector and introduces additional supervision only during training, without modifying the inference architecture, thus incurring no extra inference cost.

\textbf{Domain Augmentation.}
A practical route for single-domain generalization is to synthesize multiple `pseudo-domains' from the source domain during training while keeping inference unchanged. Policy-based methods, such as SRA~\cite{xiao2025sample}, adapt augmentation strength to sample difficulty but remain limited by predefined transformation spaces. Learning-based methods further expand the source-domain distribution: ABA~\cite{cheng2023adversarial} learns stochastic augmentation policies, MAD~\cite{xu2023multi} suppresses non-causal factors via multi-view adversarial learning, and NP~\cite{fan2023towards} perturbs global feature statistics, though synthesized variations are often hard to control. Structure-aware methods, such as OA-DG~\cite{lee2024object}, preserve spatial consistency via object-aware mixing but depend on the realism of synthesized domains. PhysAug~\cite{xu2025physaug} introduces frequency-space perturbations to simulate imaging changes, yet the induced style variations remain limited. 
Despite their effectiveness, representations learned from augmented domains often generalize poorly beyond the predefined augmentation space, and \textbf{their design focus is not on the object-level discriminative issues of detectors under domain shift}. In contrast, our method operates only during training and focuses on enforcing the diversity of object-level discriminative evidence and its consistency across views, rather than relying on broader domain synthesis.

\section{The Proposed Method} 
\subsection{Preliminaries}
\textbf{Setting.}
Let $D_s$ be the single source domain containing $N^s$ labelled training examples 
$\{(x_i^s, y_i^s)\}_{i=1}^{N^s}$, where $x_i \in \mathbb{R}^{H \times W \times C}$ is an image 
and $y_i = \{k_i, b_i\}$ is the corresponding label with bounding box coordinates 
$b_i \in \mathbb{R}^4$ and the associated category label $k_i \in \{1, \dots, K\}$. 
$K$ is the number of categories, and $H$, $W$, and $C$ represent the height, width, 
and number of channels, respectively.
Let $\{D_t\}_{t=1}^{T}$ be the set of $T$ (unseen) target domains. Our goal is to learn an object 
detector using training examples from $D_s$ that generalizes well on test examples from $D_t$. 
We assume that both $D_s$ and $D_t$ share the same label space.
Unlike conventional SDGOD, RT-SDGOD imposes stricter real-time constraints: no extra modules, parameter updates, adaptation, or computation are allowed at inference. Thus, cross-domain robustness must come solely from training-time learning, with no extra inference overhead beyond the base real-time detector.

\textbf{O2M supervision in DETR-based detectors.}
At decoder layer $l$, let $\mathbf{F}^{(l)} \in \mathbb{R}^{Q \times d}$ denote the object query features, where $Q$ is the number of queries and $d$ is the feature dimension. The classification and regression heads produce
$\mathbf{C}^{(l)} = f_{\mathrm{cls}}(\mathbf{F}^{(l)})$ and $\mathbf{B}^{(l)} = f_{\mathrm{box}}(\mathbf{F}^{(l)})$,
where $\mathbf{C}^{(l)} \in \mathbb{R}^{Q \times K}$ are the classification logits over $K$ categories, and $\mathbf{B}^{(l)} \in \mathbb{R}^{Q \times 4}$ are the predicted boxes.

Standard DETR~\cite{carion2020end} adopts one-to-one (O2O) matching, assigning each ground-truth (GT) object to a single query, whereas O2M supervision~\cite{li2022dn,jia2023detrs,chen2023group,zhao2024ms} matches multiple queries to the same object. For the $n$-th object, the matched queries form an object-specific group $\mathcal{G}_n^{(l)} = \{ q_{n,1}^{(l)}, \dots, q_{n,S_n}^{(l)} \}$,
where $q_{n,i}^{(l)}$ is the $i$-th matched query and $S_n$ is the group size. All queries in $\mathcal{G}_n^{(l)}$ share the same supervision target. The O2M loss is
\begin{equation}
\scriptsize
\mathcal{L}_{\mathrm{O2M}}^{(l)}=\sum_{n=1}^{N}\sum_{i=1}^{S_n}
\left[\ell_{\mathrm{cls}}(k_{n,i}^{(l)}, \bar{k}_n)+\ell_{\mathrm{box}}(b_{n,i}^{(l)}, \bar{b}_n)\right],
\end{equation}
where $N$ is the number of GT objects, $k_{n,i}^{(l)}$, $b_{n,i}^{(l)}$ denote the classification and box predictions of $q_{n,i}^{(l)}$, $(\bar{k}_n,\bar{b}_n)$ are the label and box annotation of the $n$-th object.

\subsection{Overview}
\begin{figure}[t!]
	\centering
	\includegraphics[width=1.0\linewidth]{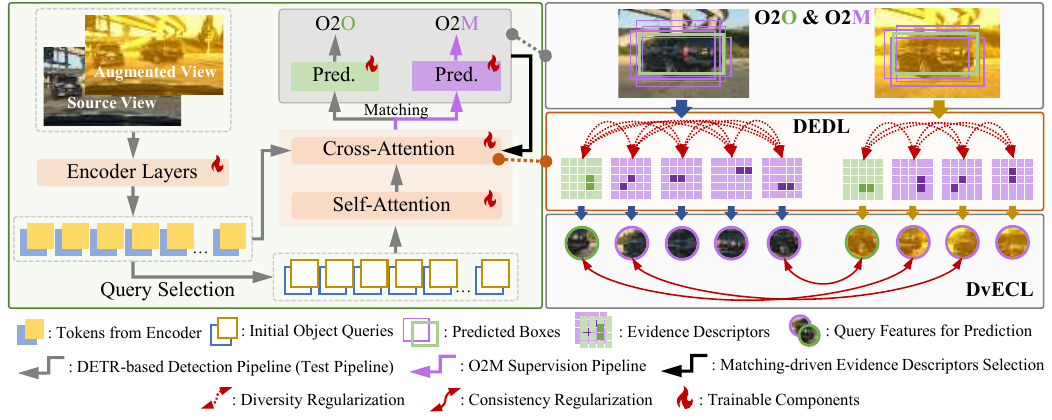}
	\vspace{-20pt}
    \caption{Overview of RT-SDGDet. Built on RF-DETR with O2M supervision, RT-SDGDet introduces Discriminative Evidence Diversity Learning (DEDL) and Dual-view Evidence Consistency Learning (DvECL) during training, without changing the inference pipeline.}
	\label{fig:overall}
	\vspace{-15pt}
\end{figure}

As shown in Fig.~\ref{fig:overall}, built on the DETR-based real-time detector RF-DETR~\cite{robinson2025rf} and inspired by the O2M supervision mechanism in MS-DETR~\cite{zhao2024ms}, we propose RT-SDGDet, a multi-evidence collaborative modeling framework for RT-SDGOD. Specifically, we first leverage O2M supervision to assign multiple queries to each target and refine the assignment process to construct more stable object-specific query groups. 
Then, at each decoder layer, we construct an explicit evidence descriptor for each query based on deformable cross-attention, and introduce \textbf{Discriminative Evidence Diversity Learning (DEDL)} to encourage queries within the same group to focus on complementary discriminative evidence via diversity regularization, thereby expanding object-level evidence coverage.
Furthermore, between the original image and its augmented view, we establish cross-view query correspondences based on evidence descriptor similarity and apply \textbf{Dual-view Evidence Consistency Learning (DvECL)} to enforce consistency between query features relying on similar discriminative evidence, improving representation stability under view perturbations. O2M, DEDL, and DvECL are introduced only during training without changing the inference structure, and thus incur no extra inference cost.

\subsection{Discriminative Evidence Diversity Learning (DEDL)}
Existing DETR-based real-time detectors often rely on limited and concentrated discriminative evidence, which becomes fragile under domain shifts once such evidence degrades. We therefore introduce DEDL to encourage more sufficient and diverse object-level evidence.

\textbf{O2M-based query group construction.}
O2M supervision assigns multiple queries to each target, enabling diversified discriminative evidence learning for the same object. However, to support more stable multi-evidence learning, we refine the assignment procedure while preserving the original O2M supervision form, yielding more stable and exclusive object-specific query groups.

Specifically, at decoder layer $l$, we compute a query--target matching quality matrix $\mathbf{T}^{(l)} \in \mathbb{R}^{N \times Q}$ following prior O2M matching, where $N$ and $Q$ are the numbers of GT objects and queries. Each entry is
\begin{equation}
\label{eq:o2m}
\scriptsize
T_{nq}^{(l)}=\lambda_{\mathrm{box}}\,\mathrm{IoU}(b_q^{(l)},\bar{b}_n)+\lambda_{\mathrm{cls}}\,p_q^{(l)}(\bar{k}_n),
\end{equation}
where $b_q^{(l)}$ is the box predicted by query $q$ at layer $l$, and $(\bar{b}_n,\bar{k}_n)$ are the GT box and label of the $n$-th object. $p_q^{(l)}(\bar{k}_n)$ denotes the confidence for category $\bar{k}_n$, and $\lambda_{\mathrm{box}}$ and $\lambda_{\mathrm{cls}}$ are balancing coefficients.
We then refine the assignment by (i) allocating each object one unused query to ensure initial coverage, and (ii) greedily adding high-quality queries subject to exclusive assignment and at most $W$ queries per object, yielding object-specific query groups $\{\mathcal{G}_n^{(l)}\}$. The O2O-matched query is also merged into each group.

\textbf{Query-level evidence descriptor.}
To explicitly model discriminative evidence within each group, we construct a query-level evidence descriptor from deformable cross-attention. Unlike query features that entangle semantics, localization, and context, deformable cross-attention directly indicates where a query gathers evidence and how strongly it relies on each sampled region. We therefore use the attention weights together with their sampling locations as an explicit evidence descriptor for each query.

At decoder layer $l$, let $\mathbf{W}^{(l)} \in \mathbb{R}^{B \times Q \times H \times LP}$ and $\mathbf{S}^{(l)} \in \mathbb{R}^{B \times Q \times H \times L \times P \times 2}$ denote the deformable cross-attention weights and sampling locations, where $B$ is the batch size, $Q$ is the number of queries, $H$ is the number of attention heads, $L$ is the number of feature levels, and $P$ is the number of sampling points per level. We first reshape $\mathbf{W}^{(l)}$ into $\mathbb{R}^{B \times Q \times H \times L \times P}$ and average both tensors over heads to obtain query-level attention weights and sampling coordinates. Let $\mathbf{w}^{(l)} \in \mathbb{R}^{B \times Q \times LP}$ denote the flattened attention weights, and $\mathbf{x}^{(l)}, \mathbf{y}^{(l)} \in \mathbb{R}^{B \times Q \times LP}$ denote the corresponding flattened sampling coordinates. We then construct the evidence descriptor as
\begin{equation}
\scriptsize
\mathbf{E}^{(l)} = \mathrm{Norm}\!\left(\left[\mathbf{w}^{(l)},\; \mathbf{w}^{(l)} \odot \mathbf{x}^{(l)},\; \mathbf{w}^{(l)} \odot \mathbf{y}^{(l)}\right]\right),
\end{equation}
where $\odot$ denotes element-wise multiplication, $[\cdot,\cdot,\cdot]$ denotes concatenation along the last dimension, and $\mathrm{Norm}(\cdot)$ denotes $\ell_2$ normalization. The resulting $\mathbf{E}^{(l)} \in \mathbb{R}^{B \times Q \times 3LP}$ encodes both the attention strength and the spatial source of the evidence used by each query.

\textbf{Diversity regularization.}
Given an object-specific group $\mathcal{G}_n^{(l)}$, we encourage its queries to capture complementary discriminative evidence by penalizing excessive similarity between their evidence descriptors.
Let $\mathcal{E}_n^{(l)}=\{\mathbf{e}_{n,1}^{(l)},\dots,\mathbf{e}_{n,S_n}^{(l)}\}$ denote the normalized evidence descriptors of $\mathcal{G}_n^{(l)}$. The layer-wise diversity loss is defined as
\begin{equation}
\label{eq:m}
\scriptsize
\mathcal{L}_{\mathrm{div}}^{(l)}
=
\frac{1}{N_l}
\sum_n
\frac{1}{S_n(S_n-1)}
\sum_{i\neq j}
\max\big(0,\langle \mathbf{e}_{n,i}^{(l)},\mathbf{e}_{n,j}^{(l)}\rangle - m\big),
\end{equation}
where $N_l$ is the number of valid groups at layer $l$, $m$ is a similarity margin, and $\langle\cdot,\cdot\rangle$ denotes inner product. We apply this regularization at every decoder layer and sum the losses over all $L$ layers:
$\mathcal{L}_{\mathrm{DEDL}}=\sum_{l=1}^{L} \mathcal{L}_{\mathrm{div}}^{(l)}$.
\subsection{Dual-view Evidence Consistency Learning (DvECL)}
While DEDL promotes complementary evidence learning for the same object, such specialization may remain unstable under view perturbations. We therefore introduce DvECL to align cross-view queries of the same object that rely on similar discriminative evidence, thereby improving cross-view stability.

\textbf{Dual-view object-specific query groups.}
During training, for each source image, we construct an augmented view using MAD~\cite{xu2023multi} and feed both views into the detector jointly. Let the original and augmented views be denoted by $I^{s}$ and $\tilde{I}^{s}$, respectively. Since they originate from the same image, they share the same object identities and labels. Based on the layer-wise O2M assignment, we obtain two object-specific query groups for the $n$-th object at decoder layer $l$:
\begin{equation}
\scriptsize
\mathcal{G}_{n,\mathrm{src}}^{(l)} = \{ q_{n,1,\mathrm{src}}^{(l)}, \dots, q_{n,S_n^{\mathrm{src}},\mathrm{src}}^{(l)} \},
\qquad
\mathcal{G}_{n,\mathrm{aug}}^{(l)} = \{ q_{n,1,\mathrm{aug}}^{(l)}, \dots, q_{n,S_n^{\mathrm{aug}},\mathrm{aug}}^{(l)} \},
\end{equation}
where $\mathcal{G}_{n,\mathrm{src}}^{(l)}$ and $\mathcal{G}_{n,\mathrm{aug}}^{(l)}$ denote the $n$-th object's query groups in the original and augmented views, respectively, and $S_n^{\mathrm{src}}$ and $S_n^{\mathrm{aug}}$ are their group sizes.

\textbf{Evidence-descriptor-guided query pairing.}
Directly aligning the two query groups is inappropriate because their sizes and query ordering may differ across views. Instead, we establish cross-view correspondences based on evidence descriptor similarity, so that queries relying on similar discriminative evidence are aligned under different view perturbations.

Let $\mathbf{E}_{\mathrm{src}}^{(l)}$ and $\mathbf{E}_{\mathrm{aug}}^{(l)}$ denote the query-level evidence descriptors of the original and augmented views at layer $l$, respectively. For the $n$-th object, we compute the pairwise evidence similarity matrix
\begin{equation}
\scriptsize
\mathbf{M}_{n}^{(l)}(i,j)
=
\left\langle
\mathbf{e}_{n,i,\mathrm{src}}^{(l)},
\mathbf{e}_{n,j,\mathrm{aug}}^{(l)}
\right\rangle,
\end{equation}
where $\mathbf{e}_{n,i,\mathrm{src}}^{(l)} \in \mathcal{G}_{n,\mathrm{src}}^{(l)}$ and $\mathbf{e}_{n,j,\mathrm{aug}}^{(l)} \in \mathcal{G}_{n,\mathrm{aug}}^{(l)}$ are normalized evidence descriptors. Based on $\mathbf{M}_{n}^{(l)}$, we establish cross-view correspondences using a mutual nearest neighbor strategy. Specifically, a query pair $(i,j)$ is retained only if query $i$ takes query $j$ as its most similar counterpart, query $j$ also takes query $i$ as its most similar counterpart, and their similarity exceeds a threshold $\tau$. The resulting pairs are denoted by $\mathcal{P}^{(l)}$.

\textbf{Consistency regularization.}
Given a paired correspondence $(i,j)\in \mathcal{P}^{(l)}$, let $\mathbf{h}_{n,i,\mathrm{src}}^{(l)}$ and $\mathbf{h}_{n,j,\mathrm{aug}}^{(l)}$ denote the corresponding query features. We enforce consistency on these evidence-aligned query features using cosine distance:
\begin{equation}
\scriptsize
\mathcal{L}_{\mathrm{cons}}^{(l)}
=
\frac{1}{M_l}
\sum_{(i,j)\in \mathcal{P}^{(l)}}
\ell_{\mathrm{cons}}
\left(
\mathbf{h}_{n,i,\mathrm{src}}^{(l)},
\mathbf{h}_{n,j,\mathrm{aug}}^{(l)}
\right),
\end{equation}
where $M_l$ is the number of valid query pairs at layer $l$, and $\ell_{\mathrm{cons}}(\cdot,\cdot)$ is the cosine distance loss. The final consistency loss is obtained by summing over all $L$ decoder layers: $\mathcal{L}_{\mathrm{DvECL}}=\sum_{l=1}^{L}\mathcal{L}_{\mathrm{cons}}^{(l)}$.

\subsection{Training Objective and Implementation Details}
\textbf{Training objective.} Our training loss consists of four components: the standard O2O DETR-based detection loss $\mathcal{L}_{\mathrm{O2O}}$, the O2M detection loss $\mathcal{L}_{\mathrm{O2M}}$, the proposed discriminative evidence diversity loss $\mathcal{L}_{\mathrm{DEDL}}$, and the proposed dual-view evidence-guided consistency loss $\mathcal{L}_{\mathrm{DvECL}}$. The total loss is formulated as
\begin{equation}
\label{eq:overall_loss}
\scriptsize
\mathcal{L}=\mathcal{L}_{\mathrm{O2O}}+\lambda_{\mathrm{O2M}} \mathcal{L}_{\mathrm{O2M}}+\lambda_{\mathrm{DEDL}} \mathcal{L}_{\mathrm{DEDL}}+\lambda_{\mathrm{DvECL}} \mathcal{L}_{\mathrm{DvECL}},
\end{equation}
where $\lambda_{\mathrm{O2M}}$, $\lambda_{\mathrm{DEDL}}$, and $\lambda_{\mathrm{DvECL}}$ are balancing coefficients.

\textbf{Implementation details.}
We adopt RF-DETR as the baseline and conduct all experiments on 8×3090 GPUs under distributed data parallel training. The model is trained for 48 epochs with a batch size of 2 per GPU and 8-step gradient accumulation. We use AdamW with base learning rates of $1\times10^{-4}$ for the detector and $1.5\times10^{-4}$ for the encoder, and a weight decay of $1\times10^{-4}$. The learning rate is decayed at epoch 11, and gradient clipping with a maximum norm of 0.1 is applied to stabilize training.
For the O2M setting in Eq.~(\ref{eq:o2m}), we follow MS-DETR and set $\lambda_{\mathrm{box}}=0.7$, $\lambda_{\mathrm{cls}}=0.3$, and the maximum number of retained queries per object to 6. In Eq.~(\ref{eq:m}), we set the similarity margin $m$ to 0.3, and the evidence descriptor similarity threshold $\tau$ to 0.8. In Eq.~(\ref{eq:overall_loss}), the loss weights $\lambda_{\mathrm{O2M}}$, $\lambda_{\mathrm{DEDL}}$, and $\lambda_{\mathrm{DvECL}}$ are set to 0.5, 0.5, and 0.3, respectively. We further analyze these hyperparameters in the ablation study.

\section{Experimental Results}

\subsection{Setup}

\textbf{Datasets.} We conduct experiments on the SDGOD benchmark~\cite{wu2022single}, which contains 5 weather conditions: Daytime-Clear, Daytime-Foggy, Dusk-Rainy, Night-Clear, and Night-Rainy. Daytime-Clear is used as the source domain, with 19,395 images for training and 8,313 for testing. The remaining four conditions serve as target domains, including 3,775 images in Daytime-Foggy, 3,501 in Dusk-Rainy, 26,158 in Night-Clear, and 2,494 in Night-Rainy. The dataset provides annotations for seven object categories: person, car, bike, rider, motor, bus, and truck.

\textbf{Evaluation metric.}
For evaluation, we adopt mean Average Precision (mAP) as the primary performance metric and follow the standard evaluation protocol for SDGOD, reporting all results at an IoU threshold of 50\% (mAP@50).

\textbf{Comparison methods.}
Since RT-SDGOD requires a fixed test-time inference pipeline under real-time constraints, most existing SDGOD methods are not suitable for fair comparison, as they are typically built on Faster R-CNN or rely on extra auxiliary modules and test-time mechanisms. Therefore, we consider two types of baselines: recent representative real-time detectors, and domain augmentation methods used only during training without changing the test-time inference structure.
Specifically, the real-time detector baselines include YOLOv12~\cite{tian2025yolov12}, YOLOv13~\cite{lei2025yolov13}, LW-DETR~\cite{chen2024lw}, D-FINE~\cite{peng2024d},  as well as  DEIM~\cite{huang2025deim}, DEIMv2~\cite{huang2025real}, RT-DETRv3~\cite{wang2025rt}, RT-DETRv4~\cite{liao2025rt}, and RF-DETR~\cite{robinson2025rf}. 
In the subsequent experiments and analyses, to ensure a fair comparison, we uniformly adopt the variant of each method whose model scale is closest to that of the main baseline, RF-DETR-L, \textbf{and the results for other model scales are provided in the supplementary material}.
For domain augmentation baselines, we consider ABA~\cite{cheng2023adversarial}, NP~\cite{fan2023towards}, MAD~\cite{xu2023multi}, OA-DG~\cite{lee2024object}, SRA~\cite{xiao2025sample}, and PhysAug~\cite{xu2025physaug}. To ensure a fair comparison under strict real-time constraints, we uniformly apply all these methods to the baseline RF-DETR-L for evaluation.



\begin{table}[t!]
\centering
\tiny
\setlength{\tabcolsep}{1.8pt}
\caption{Generalization Results on Five Different Domains (\%).} \vspace{-10pt}
\begin{tabular}{l|ccccc>{\columncolor{gray!15}}c|cc}
\hline
Methods & D-Clear & N-Clear & D-Foggy & D-Rainy & N-Rainy & Avg.$\uparrow$ & Params(M) & GFLOPs$\downarrow$ \\
\hline
YOLOv12-L &62.7 &48.1 &43.0 &40.1 &22.4 &43.3 &26.4 &88.9  \\
YOLOv13-L &61.8 &47.3 &42.2 &40.4 &23.1 &43.0 &27.6 &88.4  \\
LW-DETR-M &61.1 &49.1 &42.0 &47.1 &29.0 &45.7 &28.2 &42.8  \\
D-FINE-L &63.1 &49.4 &40.0 &43.2 &23.2 &43.8 &31.0 &91.0  \\
DEIM-L &64.1 &50.3 &43.0 &43.6 &24.8 &45.2 &31.0 &91.0  \\
DEIMv2-L &63.9 &52.1 &43.4 &51.0 &32.9 &48.7 &32.0 &96.0  \\
RT-DETRv3-R34 &62.8 &46.7 &40.1 &39.1 &16.9 &41.1 &31.0 &92.0  \\
RT-DETRv4-L &65.9 &52.2 &43.5 &47.7 &27.5 &47.4 &31.0 &91.0  \\
\hline
RF-DETR-L &68.6 &56.4 &48.2 &52.6 &33.5 &51.9 &33.9 &125.6  \\
\hline
~~+ABA &68.0 &56.7 &48.3 &55.0 &37.6 &53.1 &33.9 &125.6  \\
~~+NP  &68.8&57.2&48.6&55.0&36.5&53.2&33.9 &125.6   \\                                                                       
~~+MAD &68.7 &57.1 &48.9 &54.7 &36.9 &53.3 &33.9 &125.6  \\
~~+OA-DG &68.3 &56.7 &48.6 &54.8 &34.1 &52.5 &33.9 &125.6  \\
~~+SRA &68.3 &56.2 &\textbf{49.9} &54.7 &36.0 &53.0 &33.9 &125.6  \\ 
~~+PhysAug &68.4 &56.9 &48.1 &55.2 &37.4 &53.2 &33.9 &125.6  \\
\hline
RT-SDGDet &\textbf{68.9}&\textbf{58.0}&\textbf{49.9}&\textbf{56.2}&\textbf{39.0}&\textbf{54.4} &33.9 &125.6 \\
\hline
\end{tabular}\vspace{-15pt}
\label{tab:avg}
\end{table}

\subsection{Main Results}

\textbf{Overall results across all test domains.}
As shown in Table~\ref{tab:avg}, RT-SDGDet delivers consistent gains across all five domains and achieves the best overall performance with an average mAP of 54.4, outperforming the RF-DETR-L baseline by 2.5 mAP. Moreover, RT-SDGDet ranks first on all four unseen target domains, with the largest gains on the more severely degraded Night-Rainy and Dusk-Rainy scenes (+5.5 and +3.6 mAP, respectively), while still improving Daytime-Foggy and Night-Clear, where the domain shift is relatively milder, by +1.7 and +1.6 mAP. This trend is consistent with our motivation, suggesting that learning more sufficient and stable object-level discriminative evidence is particularly important under severe degradation. Meanwhile, RT-SDGDet preserves the same test-time parameter count and GFLOPs as the RF-DETR-L baseline, showing that these gains come at zero extra inference cost.

Compared with domain augmentation methods, RT-SDGDet also achieves superior overall performance. While MAD, NP, and PhysAug improve some domains, RT-SDGDet performs better in both average mAP and the most challenging ones. This suggests that domain expansion alone is insufficient to mitigate domain-shift degradation in real-time detectors, whereas explicitly improving the sufficiency and stability of object-level discriminative evidence is more effective.

\begin{table}[t!]
\centering
\tiny 
\caption{The results on the \textbf{Night-Clear} and  \textbf{Night-Rainy} scenes (\%).} \vspace{-10pt}
\setlength{\tabcolsep}{1.3pt}
\begin{tabular}{l|ccccccc>{\columncolor{gray!15}}c|ccccccc>{\columncolor{gray!15}}c}
\hline
\multirow{2}{*}{Methods} 
& \multicolumn{8}{c|}{Night-Clear} 
& \multicolumn{8}{c}{Night-Rainy} \\
\cline{2-17}
  & Bus & Bike & Car & Motor & Person & Rider & Truck & mAP  & Bus & Bike & Car & Motor & Person & Rider & Truck & mAP\\
\hline
YOLOv12-L &48.2 &43.1 &72.8 &24.7 &55.9 &39.9 &51.8 &48.1    &36.2 &7.03 &52.4 &2.05 &17.8 &7.54 &33.6 &22.4\\
YOLOv13-L &47.9 &42.2 &72.5 &23.1 &55.6 &37.9 &51.9 &47.3      &35.2 &6.85 &54.5 &3.30 &20.6 &7.56 &33.9 &23.1\\
LW-DETR-M &53.4 &49.9 &68.2 &28.8 &51.8 &37.0 &54.8 &49.1     &43.6 &17.7 &50.3 &8.61 &27.1 &17.7 &38.0 &29.0\\
D-FINE-L &52.3 &48.4 &73.8 &24.6 &54.1 &38.9 &53.9 &49.4      &38.5 &10.6 &49.6 &4.10 &20.0 &9.64 &30.3 &23.2 \\
DEIM-L &53.7 &48.0 &74.4& 25.6 &54.7 &40.2 &55.3 &50.3     &36.1 &13.8 &52.1 &5.71 &20.1 &11.8 &34.1 &24.8\\
DEIMv2-L &58.0 &49.1 &73.9 &28.2 &54.5 &43.0 &57.8 &52.1     &52.7 &15.6 &60.5 &9.15 &29.2 &19.1 &44.2 &32.9\\
\hline
~~+ABA &53.6&41.6&58.8&27.3&49.2&39.3&55.4&47.9  &52.4&19.9&58.3&48.6&29.2&28.6&47.0&40.6   \\
~~+NP  &57.2&48.1&73.8&34.8&53.4&40.7&58.5&52.4  &54.0&15.8&61.9&48.2&28.8&27.3&47.4&40.5 \\
~~+Mad &58.0&48.9&74.0&35.2&53.5&42.5&58.9&53.0  &52.8&20.6&62.6&48.6&30.4&26.4&46.6&41.1 \\
~~+OA-DG &56.6&48.4&73.9&33.6&53.2&41.5&56.9&52.0  &52.4&17.9&61.8&47.2&28.6&24.5&46.8&39.9 \\
~~+SRA &57.3&49.4&73.7&33.8&53.7&42.6&58.3&52.7  &53.2&16.4&60.9&46.3&30.1&25.8&45.1&39.7 \\
~~+PhysAug &57.0&48.4&73.6&34.2&53.2&42.8&58.7&52.5  &55.2&24.6&62.9&50.1&31.8&29.2&51.3&43.6 \\
\hline
RT-DETRv3-R34 &50.6 &42.8 &72.7 &21.5 &52.9 &34.7 &51.4 &46.7     &30.3 &6.2 &40.9 &0.20 &10.7 &7.60 &22.3 &16.9\\
RT-DETRv4-L &55.9 &49.7 &75.6 &28.6 &56.5 &42.5 &56.9 &52.2     &44.3 &10.4 &57.2 &4.80 &23.7 &12.4 &39.5 &27.5 \\
\hline
RF-DETR-L &57.0 &55.8 &75.5 &36.5 &63.1 &47.5 &59.6 &56.4      &44.7 &18.8 &60.0 &9.90 &37.3 &19.2 &44.5 &33.5 \\
\hline
~~+ABA &\textbf{60.7} &56.1 &74.6 &\textbf{37.2} &62.0 &46.1 &60.2 &56.7     &52.3 &22.4 &61.9 &18.2 &37.6 &20.9 &49.6 &37.6 \\
~~+NP  &59.3&57.0&75.8&36.3&63.4&47.1&61.5&57.2    &52.4&21.8&62.3&11.7&38.2&20.2&49.1&36.5  \\
~~+MAD &59.6 &\textbf{57.5} &74.7 &37.1 &62.5 &47.2 &61.4 &57.1      &50.6 &20.9 &\textbf{62.6} &18.3 &39.3 &17.9 &48.5 &36.9 \\
~~+OA-DG &60.4 &56.0 &75.2 &35.0 &62.0 &47.2 &61.3 &56.7      &48.0 &19.9 &59.4 &10.3 &35.9 &17.5 &47.4 &34.1 \\
~~+SRA &60.1 &56.1 &73.0 &35.0 &61.1 &47.0 &61.0 &56.2       &52.1 &19.2 &61.2 &14.4 &37.8 &19.7 &47.4 &36.0 \\
~~+PhysAug &60.5 &57.1 &74.2 &37.0 &62.1 &46.1 &61.3 &56.9     &\textbf{54.0} &22.6 &59.3 &14.7 &40.0 &21.1 &50.1 &37.4  \\
\hline
RT-SDGDet &60.3&57.0&\textbf{75.9}&37.1&\textbf{64.4}&\textbf{48.8}&\textbf{62.3}&\textbf{58.0}      &53.8&\textbf{22.9}&61.8&\textbf{20.0}&\textbf{41.3}&\textbf{22.7}&\textbf{50.5}&\textbf{39.0}   \\
\hline
\end{tabular}\vspace{-5pt}
\label{tab:2}
\end{table}

\begin{table}[t!]
\centering
\tiny 
\caption{The results on the \textbf{Dusk-Rainy} and \textbf{Daytime-Foggy} scenes (\%).}\vspace{-10pt}
\setlength{\tabcolsep}{1.3pt}
\begin{tabular}{l|ccccccc>{\columncolor{gray!15}}c|ccccccc>{\columncolor{gray!15}}c}
\hline
\multirow{2}{*}{Methods} 
& \multicolumn{8}{c|}{Dusk-Rainy} 
& \multicolumn{8}{c}{Daytime-Foggy} \\
\cline{2-17}
 & Bus & Bike & Car & Motor & Person & Rider & Truck & mAP    & Bus & Bike & Car & Motor & Person & Rider & Truck & mAP \\
\hline
YOLOv12-L &50.4 &19.6 &76.6 &18.8 &38.1 &21.1 &55.8 &40.1       &40.0 &33.3 &68.1 &35.7 &46.6 &46.1 &31.1 &43.0\\
YOLOv13-L &51.3 &19.6 &77.3 &17.8 &39.3 &19.7 &57.5 &40.4      &40.2 &31.4 &67.6 &36.1 &44.9 &44.7 &30.4 &42.2\\
LW-DETR-M &52.9 &39.0 &74.0 &30.5 &45.0 &30.2 &58.3 &47.1      &40.2 &33.3 &61.9 &39.7 &41.9 &42.4 &34.3 &42.0\\
D-FINE-L &48.8 &31.0 &76.7 &23.3 &41.8 &26.5 &54.7 &43.2      &36.5 &30.6 &64.9 &34.3 &41.7 &42.2 &29.7 &40.0 \\
DEIM-L &50.9 &31.0 &77.8 &24.0 &40.6 &23.7 &57.5 &43.6     &37.8 &33.2 &69.2 &36.9 &45.4 &45.4 &33.0 &43.0\\
DEIMv2-L &57.0 &34.1 &79.9 &47.5 &47.5 &29.1 &62.2 &51.0     &41.5 &31.4 &67.4 &39.7 &43.6 &44.3 &36.2 &43.4\\
\hline
~~+ABA &52.8&30.2&78.0&62.2&44.9&26.3&60.1&50.6  &40.6&26.5&66.6&31.5&40.4&40.1&35.0&40.1   \\
~~+NP  &58.4&34.2&80.6&65.7&48.5&29.8&63.8&54.5  &40.5&29.7&67.7&37.5&42.8&42.4&35.6&42.3 \\
~~+Mad &57.1&34.1&80.5&65.1&47.9&28.5&62.9&53.7  &42.9&32.2&70.1&39.3&45.0&45.8&37.3&44.6 \\
~~+OA-DG &58.4&33.3&80.4&65.2&48.5&26.7&63.2&53.7  &40.5&30.8&67.8&39.0&43.4&43.3&36.6&43.1 \\
~~+SRA &57.5&33.4&80.6&65.3&48.6&28.5&62.6&53.8  &43.4&31.3&69.6&38.0&44.8&44.5&37.6&44.2 \\
~~+PhysAug &58.3&34.5&80.7&65.6&49.8&29.1&64.1&54.6  &41.6&31.0&68.7&37.9&43.6&43.9&36.4&43.3 \\
\hline
RT-DETRv3-R34 &47.3 &23.1 &75.8 &18.2 &36.7 &18.9 &53.5 &39.1     &37.1 &29.8 &65.9 &32.5 &43.0 &42.6 &30.0 &40.1\\
RT-DETRv4-L &55.0 &29.3 &79.8 &41.6 &44.3 &23.7 &60.3 &47.7     &33.1 &39.6 &68.7 &39.1 &44.9 &44.9 &33.9 &43.5 \\
\hline
RF-DETR-L &58.5 &42.8 &81.3 &33.5 &55.4 &32.3 &64.6 &52.6     &49.3 &37.0 &68.2 &44.7 &49.1 &49.2 &40.0 &48.2 \\
\hline
~~+ABA &\textbf{62.1} &44.0 &80.1 &38.4 &56.0 &37.2 &67.0 &55.0     &48.2&36.2&67.8&44.1&50.2&50.1&41.4&48.3 \\
~~+NP   &61.6&47.3&\textbf{82.1}&36.2&57.0&33.9&66.7&55.0    &\textbf{49.5}&37.9&68.9&44.5&49.6&50.2&39.8&48.6 \\
~~+MAD &60.0 &45.7 &81.7 &38.1 &55.9 &36.4 &65.2 &54.7     &49.2 &37.6 &69.1 &44.0 &50.0 &50.1 &42.0 &48.9  \\
~~+OA-DG &61.2 &47.4 &81.4 &\textbf{39.3} &54.9 &33.5 &66.1 &54.8     &49.0 &38.0 &68.7 &45.2 &49.6 &49.2 &40.6 &48.6 \\
~~+SRA &61.1 &47.1 &81.3 &37.0 &55.2 &35.1 &65.9 &54.7     &49.4 &\textbf{38.9} &\textbf{70.3} &\textbf{46.0} &\textbf{51.0} &\textbf{50.8} &43.0 &\textbf{49.9} \\
~~+PhysAug &61.1 &\textbf{48.3} &80.1 &36.1 &56.0 &37.5 &\textbf{67.4} &55.2     &48.5 &38.1 &69.3 &41.0 &50.4 &48.5 &41.0 &48.1 \\
\hline
RT-SDGDet  &61.4&48.0&81.8&38.8&\textbf{57.8}&\textbf{38.1}&67.3&\textbf{56.2}    &49.4&\textbf{38.9}&\textbf{70.3}&45.8&\textbf{51.0}&50.5&\textbf{43.6}&\textbf{49.9}\\
\hline
\end{tabular}\vspace{-15pt}
\label{tab:3}
\end{table}

\textbf{Cross-domain detection results analysis.}
In Tables~\ref{tab:2} and \ref{tab:3}, we report detailed results on the four unseen target domains. Overall, RT-SDGDet achieves the best performance on all target domains. From the class-wise breakdown, the gains are not confined to a few categories, but exhibit consistent patterns aligned with different degradation types.
On Night-Clear, improvements are relatively balanced: Bus, Truck, Person, and Rider increase by +3.3, +2.7, +1.3, and +1.3 mAP over the baseline, respectively, indicating stable gains under mild nighttime shift. On the more challenging Night-Rainy domain, the largest gains appear on Bus (+9.1), Motor (+10.1), and Truck (+6.0), with additional improvements on Person (+4.0) and Rider (+3.5), suggesting better preservation of effective object representations under strong noise, reflections, and low illumination, thus mitigating severe missed detections. Similarly, on Dusk-Rainy, Bike, Motor, and Rider improve most (+5.2, +5.3, and +5.8), while Bus, Person, and Truck also gain (+2.9, +2.4, +2.7), demonstrating stronger adaptation to compounded illumination changes and rainfall. On Daytime-Foggy, all categories improve, with larger gains on Truck (+3.6), Car (+2.1), Bike (+1.9), and Person (+1.9), indicating improved use of structural cues under low contrast and blurred contours.

Overall, these results show that RT-SDGDet improves not only overall mAP but also class-wise performance, especially on key categories under severe degradation. This aligns with our analysis that enhancing the sufficiency and stability of object-level discriminative evidence helps alleviate omission-dominated degradation under domain shift.




\begin{wraptable}{r}{0.55\textwidth}  \vspace{-34pt}
\centering
\tiny
\caption{Results on False Negative Rate (\%).}
\setlength{\tabcolsep}{1.3pt}
\begin{tabular}{l|ccccc|>{\columncolor{gray!15}}c}
\hline
Method & D-Clear & N-Clear & D-Foggy & D-Rainy & N-Rainy & Avg. \\
\hline
RF-DETR-L &8.6&11.9&28.6&16.9&26.7 &18.5 \\ \hline
RT-SDGDet &8.2&10.6&24.5&13.8&21.6 &15.7 \\
\hline
\end{tabular}\vspace{-25pt}
\label{tab:FN}
\end{wraptable}
Table~\ref{tab:FN} compares RT-SDGDet and RF-DETR in False Negative Rate across all test scenarios. RT-SDGDet reduces the miss rate in all five scenarios, further showing its effectiveness in alleviating omission-dominated degradation under domain shift.

\subsection{Ablation Study}
We conduct ablations across all test scenarios to assess each RT-SDGDet component and key hyper-parameter sensitivity.
\begin{table}[h!] \vspace{-15pt}
\centering
\tiny
\caption{Ablation study of different components (\%).}\vspace{-10pt}
\setlength{\tabcolsep}{1.3pt}
\begin{tabular}{l|ccccc|>{\columncolor{gray!15}}c}
\hline
Methods & D-Clear & N-Clear & D-Foggy & D-Rainy & N-Rainy & Avg. \\
\hline
RF-DETR-L &68.6 &56.4 &48.2 &52.6 &33.5 &51.9  \\
\hline
~~+O2M  &68.6&56.7&48.7&53.9&34.4&52.5  \\
~~+O2M+DEDL  &68.9&57.5&49.5&54.9&38.2&53.8  \\
~~+O2M+DEDL+DvECL (RT-SDGDet)  &68.9&58.0&49.9&56.2&39.0&54.4 \\
\hline
\end{tabular}\vspace{-15pt}
\label{tab:components}
\end{table}

\textbf{Different component ablations.}
Table~\ref{tab:components} shows that all modules in RT-SDGDet bring stable and complementary gains. The baseline RF-DETR-L achieves 51.9 Avg., and adding O2M improves it to 52.5, indicating that increased positive-sample supervision stabilizes object modeling and improves robustness to domain shift. Building on this, introducing DEDL further raises the average to 53.8, with larger gains under severe degradations (\eg, N-Rainy: 33.5$\rightarrow$38.2), suggesting that multiple queries learn more complementary discriminative evidence. Finally, adding DvECL boosts the average to 54.4, reaching 56.2/39.0 on D-Rainy/N-Rainy, verifying that cross-view consistency improves the stability of evidence modeling. Overall, these modules work progressively and synergistically to enhance both the sufficiency and stability of object-level evidence.

\begin{wraptable}{r}{0.55\textwidth}  \vspace{-34pt}
\centering
\tiny
\caption{Sensitivity analysis of $\lambda_{\mathrm{O2M}}$ (\%).}
\setlength{\tabcolsep}{1.3pt}
\begin{tabular}{l|ccccc|>{\columncolor{gray!15}}c}
\hline
$\lambda_{\mathrm{O2M}}$ & D-Clear & N-Clear & D-Foggy & D-Rainy & N-Rainy & Avg. \\
\hline
RF-DETR-L &68.6 &56.4 &48.2 &52.6 &33.5 &51.9 \\
\hline
0.1     &68.6&56.3&48.4&52.3&33.8&51.9\\
0.3     &68.7&56.2&47.9&53.1&33.5&51.9\\
0.5     &68.6&\textbf{56.7}&\textbf{48.7}&\textbf{53.9}&\textbf{34.4}&\textbf{52.5}\\
0.8     &\textbf{68.8}&56.4&48.5&\textbf{53.9}&34.2&52.4\\
1.0     &68.3&56.5&48.3&53.6&34.0&52.1\\
\hline
\end{tabular}\vspace{-25pt}
\label{tab:o2m}
\end{wraptable}
\textbf{Sensitivity analysis of the O2M loss weight.}
Table~\ref{tab:o2m} shows that RT-SDGDet is relatively robust to $\lambda_{\mathrm{O2M}}$, while a moderate value performs best. Small weights (0.1/0.3) bring negligible gains over RF-DETR-L (51.9 Avg.), whereas $\lambda_{\mathrm{O2M}}=0.5$ yields the best average score (52.5) with consistent improvements on degraded scenarios (\eg, D-Rainy 53.9, N-Rainy 34.4). Larger weights (0.8--1.0) slightly reduce performance (52.4/52.1), implying potential imbalance with the main detection objective. We set $\lambda_{\mathrm{O2M}}=0.5$ in the remaining experiments.

\begin{wraptable}{r}{0.50\textwidth}  \vspace{-34pt}
\centering
\tiny
\caption{Sensitivity analysis of $\lambda_{\mathrm{DEDL}}$ (\%).}
\setlength{\tabcolsep}{1.3pt}
\begin{tabular}{l|ccccc|>{\columncolor{gray!15}}c}
\hline
$\lambda_{\mathrm{DEDL}}$ & D-Clear & N-Clear & D-Foggy & D-Rainy & N-Rainy & Avg. \\
\hline
+O2M &68.6&56.7&48.7&53.9&34.4&52.5 \\
\hline
0.1  &\textbf{68.8}&56.7&48.7&54.1&34.9&52.6\\
0.3  &68.6&56.8&48.6&54.0&36.4&52.9\\
0.5  &\textbf{68.8}&\textbf{57.6}&\textbf{49.3}&\textbf{54.8}&\textbf{37.9}&\textbf{53.7}\\
0.8  &68.7&57.2&49.0&54.3&36.2&53.1\\
1.0  &68.7&57.1&48.9&53.7&35.8&52.8\\
\hline
\end{tabular}\vspace{-25pt}
\label{tab:div}
\end{wraptable}
\textbf{Sensitivity analysis of the diversity loss weight.}
We first fix the similarity margin to a reasonable value, $m=0.2$ (Table~\ref{tab:div}), and find that a moderate diversity weight works best. Starting from +O2M (52.5 Avg.), $\lambda_{\mathrm{DEDL}}=0.5$ improves the average score to 53.7, with larger gains under severe degradations (\eg, N-Rainy: 34.4$\rightarrow$37.9), indicating more complementary discriminative evidence across queries. Increasing $\lambda_{\mathrm{DEDL}}$ to 0.8--1.0 reduces performance to 53.1/52.8, suggesting that overly strong diversity regularization may hinder the main detection objective. We thus use $\lambda_{\mathrm{DEDL}}=0.5$ in the remaining experiments.

\begin{wraptable}{r}{0.50\textwidth}  \vspace{-34pt}
\centering
\tiny
\setlength{\tabcolsep}{1.3pt}
\caption{Sensitivity analysis of the $m$ (\%).}
\begin{tabular}{l|ccccc|>{\columncolor{gray!15}}c}
\hline
$m$ & D-Clear & N-Clear & D-Foggy & D-Rainy & N-Rainy & Avg. \\
\hline
+O2M    &68.6&56.7&48.7&53.9&34.4&52.5 \\
\hline
0.0     &68.4&57.0&48.7&54.2&37.2&53.1\\
0.1     &68.6&57.3&49.0&54.6&37.2&53.3\\
0.2     &68.8&\textbf{57.6}&49.3&54.8&37.9&53.7\\
0.3     &68.9&57.5&\textbf{49.5}&\textbf{54.9}&\textbf{38.2}&\textbf{53.8}\\
0.4     &68.7&\textbf{57.6}&49.4&54.8&37.8&53.7\\
0.5     &\textbf{69.0}&57.3&48.9&54.2&37.9&53.5\\
\hline
\end{tabular}\vspace{-15pt}
\label{tab:margin}
\end{wraptable}
\textbf{Sensitivity analysis of the similarity margin.}
After fixing $\lambda_{\mathrm{DEDL}}=0.5$ (Table~\ref{tab:margin}), we study the effect of the similarity margin $m$. RT-SDGDet is fairly robust to $m$, with all settings in $[0.0, 0.5]$ outperforming +O2M (52.5 Avg.). The best average result, 53.8, is achieved at $m=0.3$, with clear gains in challenging scenarios (N-Rainy 38.2, D-Rainy 54.9). This indicates that a proper margin promotes complementary evidence across queries without excessive dispersion. We therefore use $m=0.3$ in the remaining experiments.

\begin{wraptable}{r}{0.57\textwidth}  \vspace{-34pt}
\centering
\tiny
\setlength{\tabcolsep}{1.3pt}
\caption{Sensitivity analysis of $\lambda_{\mathrm{DvECL}}$ (\%).} 
\begin{tabular}{l|ccccc|>{\columncolor{gray!15}}c}
\hline
$\lambda_{\mathrm{DvECL}}$ & D-Clear & N-Clear & D-Foggy & D-Rainy & N-Rainy & Avg. \\
\hline
+O2M+DEDL    &68.9&57.5&49.5&54.9&38.2&53.8\\
\hline
0.1     &68.7&57.6&49.3&55.7&38.5&54.0\\
0.3     &\textbf{68.9}&\textbf{58.0}&\textbf{49.9}&56.2&\textbf{39.0}&\textbf{54.4}\\
0.5     &\textbf{68.9}&57.7&49.7&56.1&38.6&54.2\\
0.8     &68.7&57.7&49.6&56.0&38.3&54.1\\
1.0     &68.8&57.9&49.7&\textbf{56.4}&38.4&54.2\\
\hline
\end{tabular}\vspace{-25pt}
\label{tab:cons}
\end{wraptable}
\textbf{Sensitivity analysis of the consistency loss weight.}
Table~\ref{tab:cons} shows the effect of $\lambda_{\mathrm{DvECL}}$ on top of +O2M+DEDL. Adding DvECL improves performance over +O2M+DEDL (53.8 Avg.) across all tested weights, confirming the effectiveness of cross-view evidence consistency learning. The best average result, 54.4, is achieved at $\lambda_{\mathrm{DvECL}}=0.3$. Further increasing the weight to 0.5--1.0 slightly reduces performance to 54.1--54.2, suggesting that overly strong consistency regularization may hinder the learning of discriminative evidence diversity. We therefore set $\lambda_{\mathrm{DvECL}}=0.3$ in the remaining experiments.

\subsection{Visualization and Analysis}

\begin{figure}[t!]
	\centering
	\includegraphics[width=1.0\linewidth]{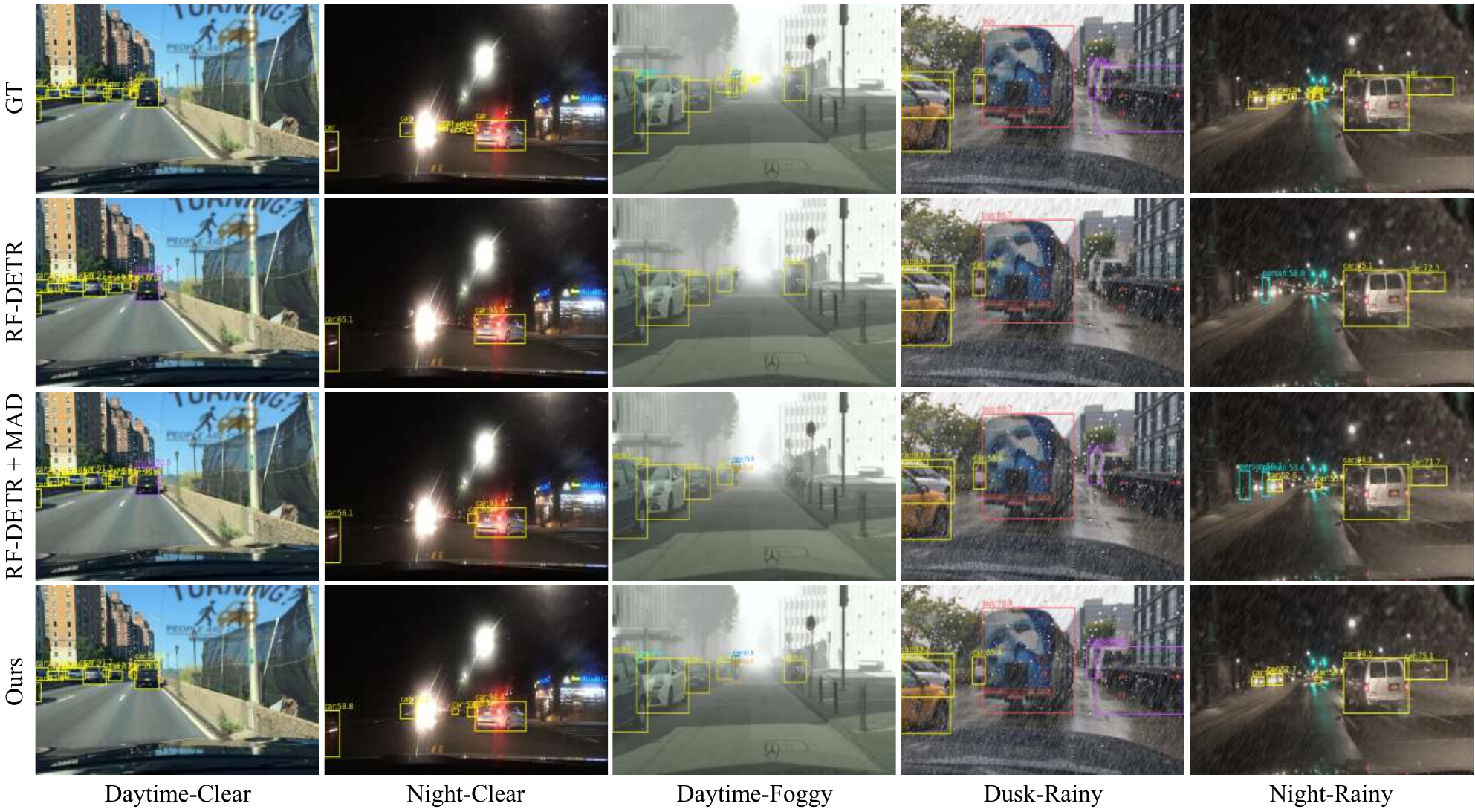}
	\vspace{-20pt}
	\caption{Qualitative comparison under different domain conditions. Comparison of detection results between RF-DETR, MAD and our method.}
	\label{fig:visual}
	\vspace{-15pt}
\end{figure}

Fig.~\ref{fig:visual} presents a qualitative comparison across the source domain and four unseen target domains. In the source-domain scene, all methods produce reasonable detections, although RF-DETR already shows occasional class confusion. As domain shift intensifies, RF-DETR degrades noticeably, mainly through missed detections on distant objects, small objects, and weather-corrupted targets. RF-DETR+MAD recovers some missed targets under degraded conditions, but still suffers from unstable localization and incomplete detections. In contrast, our method remains more stable across target domains, preserving more true detections and more complete localization, especially for distant vehicles, boundary objects, and targets degraded by fog, rain, and nighttime illumination. This is consistent with the quantitative results and supports that our method effectively mitigates omission-dominated degradation under domain shift.

\begin{wrapfigure}{r}{0.5\textwidth}\vspace{-22pt}
	\centering
	\includegraphics[width=1.0\linewidth]{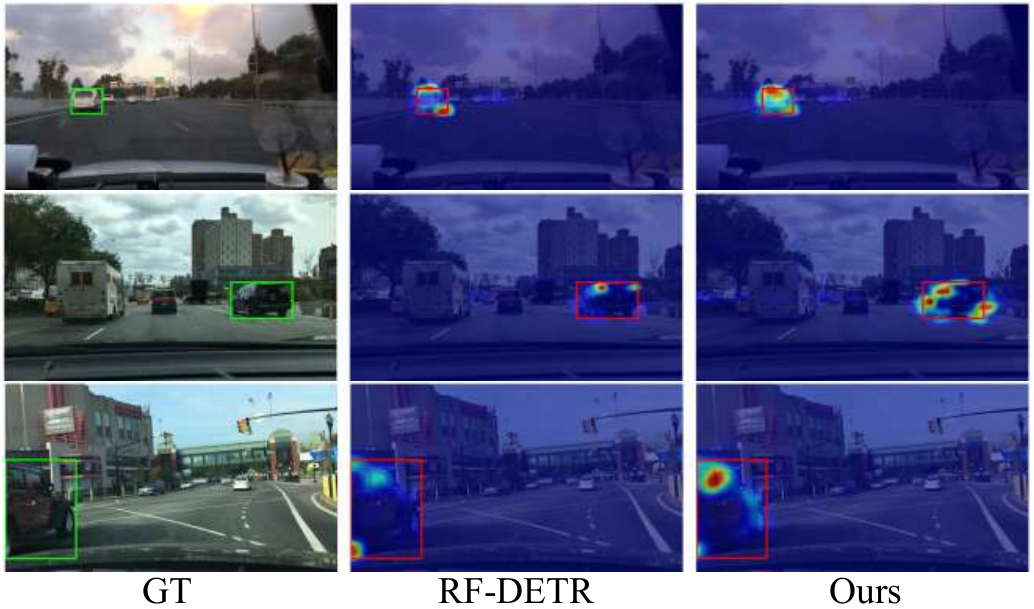}  
	\vspace{-20pt}
	\caption{Visualization of discriminative evidence regions. Compared with RF-DETR, our method yields broader and more complete responses on target objects, indicating that RT-SDGDet learns more sufficient discriminative evidence rather than relying on a few localized cues.}
	\label{fig:heatmap}
	\vspace{-20pt}
\end{wrapfigure}
To analyze the effect of our method on discriminative evidence modeling, we visualize the evidence response regions of RF-DETR and RT-SDGDet. As shown in Fig.~\ref{fig:heatmap}, the baseline RF-DETR tends to respond to only a few localized object regions, whereas our method activates broader and more complete response regions that cover more discriminative object parts. This suggests that RT-SDGDet alleviates over-reliance on fragile local cues and encourages the detector to exploit more sufficient object-level discriminative evidence.

\textbf{More Visualization and analysis are provided in the supplementary material.}

\section{Conclusion}
In this work, we formalize RT-SDGOD to study cross-domain generalization of real-time detectors under a strictly fixed inference pipeline with zero extra overhead. We find that DETR-based real-time detectors degrade under domain shift mainly due to increased missed detections, caused by limited and unstable discriminative evidence.
To address this challenge, we propose RT-SDGDet, which promotes multi-query collaboration through evidence diversity and cross-view consistency during training, improving both the sufficiency and stability of object-level evidence without changing test-time computation. Extensive experiments across multiple unseen target domains demonstrate consistent gains and strong robustness under severe weather and imaging degradations, validating the effectiveness and practicality of our framework for real-world deployment.

%
%
\bibliographystyle{splncs04_sort}
\bibliography{main}

\end{document}